\documentclass[10pt,twocolumn,letterpaper]{article}

\usepackage{cvpr}
\usepackage{times}
\usepackage{epsfig}
\usepackage{graphicx}
\usepackage{amsmath}
\usepackage{amssymb}
\usepackage{epsfig,subfigure,epstopdf,multirow}
\usepackage{booktabs}

\usepackage[pagebackref=true,breaklinks=true,letterpaper=true,colorlinks,bookmarks=false]{hyperref}

\cvprfinalcopy 

\def\cvprPaperID{3227} 

\ifcvprfinal\fi
\begin{document}

\title{Reliable and Efficient Image Cropping: A Grid Anchor based Approach}

\author{Hui Zeng$^{1}$ \qquad   Lida Li$^{1}$ \qquad Zisheng Cao$^{2}$ \qquad Lei Zhang$^{1,3}$\thanks{Corresponding author. This work is supported by HK RGC General Research Fund (PolyU 152135/16E).} \\
{\small$^{1}$The Hong Kong Polytechnic University  \qquad $^{2}$DJI Co.,Ltd  \qquad $^{3}$DAMO Academy, Alibaba Group}\\
{\tt\small \{cshzeng, cslli\}@comp.polyu.edu.hk, zisheng.cao@dji.com, cslzhang@comp.polyu.edu.hk}
}

\maketitle

\begin{abstract}
    Image cropping aims to improve the composition as well as aesthetic quality of an image by removing extraneous content from it. Existing image cropping databases provide only one or several human-annotated bounding boxes as the groundtruth, which cannot reflect the non-uniqueness and flexibility of image cropping in practice. The employed evaluation metrics such as intersection-over-union cannot reliably reflect the real performance of cropping models, either. This work revisits the problem of image cropping, and presents a grid anchor based formulation by considering the special properties and requirements (e.g., local redundancy, content preservation, aspect ratio) of image cropping. Our formulation reduces the searching space of candidate crops from millions to less than one hundred. Consequently, a grid anchor based cropping benchmark is constructed, where all crops of each image are annotated and more reliable evaluation metrics are defined. We also design an effective and lightweight network module, which simultaneously considers the region of interest and region of discard for more accurate image cropping. Our model can stably output visually pleasing crops for images of different scenes and run at a speed of 125 FPS. Code and dataset are available at: \url{https://github.com/HuiZeng/Grid-Anchor-based-Image-Cropping}.
\end{abstract}

\section{Introduction}

Cropping is an important and widely used operation to improve the aesthetic quality of captured images. It aims to remove the extraneous contents of an image, change its aspect ratio and consequently improve its composition \cite{wiki:xxx}. Since cropping is a high-frequency need in photography but a tedious job when a large number of images are to be cropped, automatic image cropping has been attracting much interest in both academia and industry in past decades \cite{chen2003visual,chor2006system,jogo2007image,yan2013learning,fang2014automatic,downing2015automated,bhatt2015multifunctional,
chen2016automatic,chen2017quantitative,wang2017deep,chedeau2017image,li2018a2}.

Early researches on image cropping mostly focused on cropping the major subject or important region of an image for small displays \cite{chen2003visual,ciocca2007self} or generating image thumbnails \cite{suh2003automatic,marchesotti2009framework}. Attention scores or saliency values were the principal concerns of these methods \cite{santella2006gaze,stentiford2007attention}. With little consideration of the overall image composition, the attention-based methods may lead to visually unpleasing outputs \cite{yan2013learning}. Moreover, user study was employed as the major criteria to subjectively evaluate cropping performance, making it very difficult to objectively compare different methods.

Recently, several benchmark databases have been released for image cropping \cite{yan2013learning,fang2014automatic,chen2017quantitative}. On these databases, one or several bounding boxes were annotated by experienced human subjects as ``groundtruth" crops for each image. Two objective metrics, namely intersection-over-union (IoU) and boundary displacement error (BDE) \cite{freixenet2002yet}, were defined to evaluate the performance of image cropping models on these databases. These public benchmarks enable many researchers to develop and test their cropping models, significantly facilitating the research on automatic image cropping \cite{yan2013learning,deng2017image,wang2017deep,chen2017quantitative,chen2017learning,deng2017aesthetic,guo2017automatic,li2018a2,wei2018good}.

\begin{figure}[t]
\centering
\subfigure{
\begin{minipage}[b]{1.0\linewidth}
\centering
\includegraphics[width=0.95\textwidth]{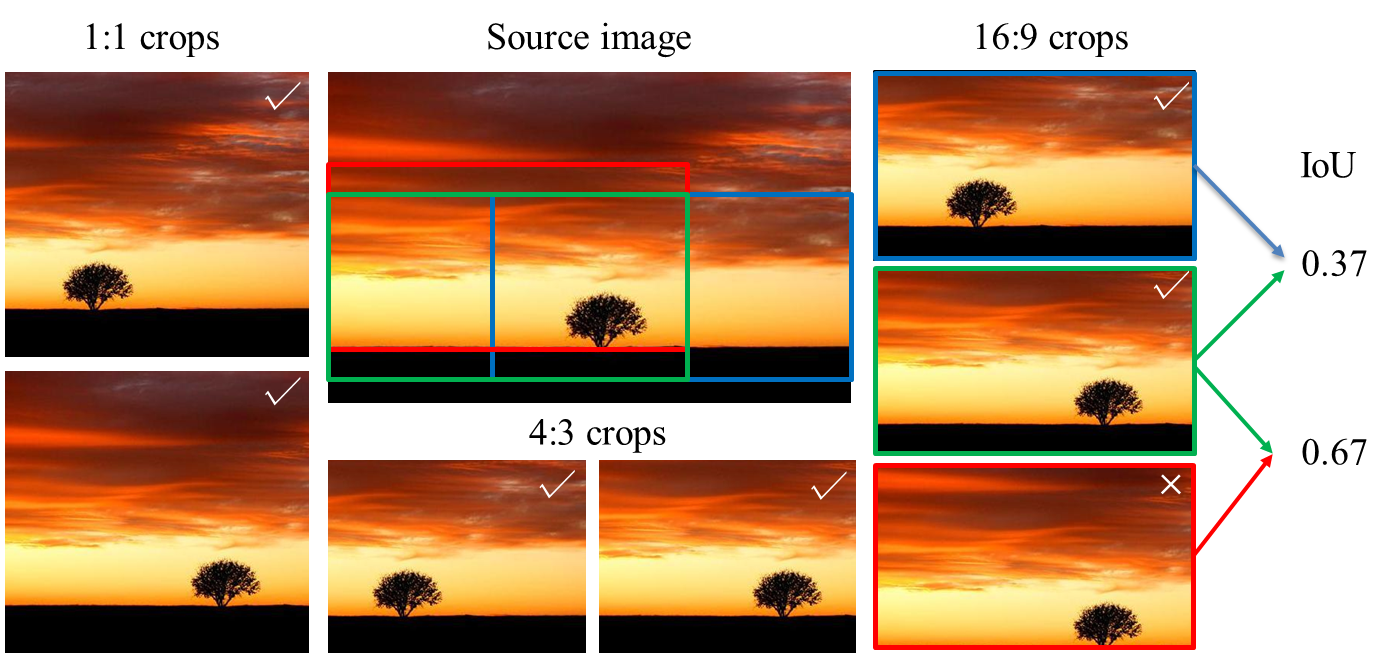}
\end{minipage}}%
\caption{The property of non-uniqueness of image cropping. Given a source image, many good crops (labeled with ``$\surd$") can be obtained under different aspect ratios (e.g., 1:1, 4:3, 16:9). Even under the same aspect ratio, there are still multiple acceptable crops. Regarding the three crops with 16:9 aspect ratio, by taking the middle one as the groundtruth, the bottom one (a bad crop, labeled with ``$\times$") will have obviously larger IoU (intersection-over-union) than the top one but with worse aesthetic quality. This shows that IoU is not a reliable metric to evaluate cropping quality.}
\label{figure:problems}
\end{figure}

Though many efforts have been made, there exists several intractable challenges caused by the special properties of image cropping.
As illustrated in Fig. \ref{figure:problems}, image cropping is naturally a subjective and flexible task without unique solution. Good crops can vary significantly under different requirements of aspect ratio and/or resolution. Even under certain aspect ratio or resolution constraint, acceptable crops can also vary.  Such a high degree of freedom makes the existing cropping databases, which have only one or several annotations, difficult to learn reliable and robust cropping models.

The commonly employed IoU or BDE metric is unreliable to evaluate the performance of image cropping models either. Referring to the three crops with 16:9 aspect ratio in Fig. \ref{figure:problems}, by taking the middle one as the groundtruth, the bottom one, which is a bad crop, will have obviously larger IoU than the top one, which is a good crop. Such a problem can be more clearly observed from Table \ref{table:Performance comparison}. By using IoU to evaluate the performance of recent works \cite{yan2013learning,wang2017deep,chen2017quantitative,chen2017learning,li2018a2} on the benchmarks ICDB \cite{yan2013learning} and FCDB \cite{chen2017quantitative}, most of them have even worse performance than the two simplest baselines: no cropping (i.e., take the source image as cropping output, denoted by Baseline\_N) or central crop (i.e., crop the central part whose width and height are 0.9 time of the source image, denoted by Baseline\_C).

\begin{table}[t]
\footnotesize
\centering
\caption{IoU scores of recent representative works on two benchmarks in comparison with two simplest baselines. Baseline\_N simply calculates the IoU between the groundtruth and source image without cropping. Baseline\_C crops the central part whose width and height are 0.9 time of the source image.}
\label{table:Performance comparison}
\begin{tabular}{|c|ccc|c|}
\hline
\multirow{2}{*}{Method} & \multicolumn{3}{c|}{ICDB\cite{yan2013learning}} & \multirow{2}{*}{FCDB\cite{chen2017quantitative}}               \\\cline{2-4}
                                            & Set 1 & Set 2 & Set 3    &                   \\\hline
Yan \etal \cite{yan2013learning}            & 0.7487 & 0.7288 & 0.7322 & --  \\
Chen \etal \cite{chen2017quantitative}      & 0.6683 & 0.6618 & 0.6483 & 0.6020  \\
Chen \etal \cite{chen2017learning}          & 0.7640 & 0.7529 & 0.7333 & \textbf{0.6802}  \\
Wang \etal \cite{wang2017deep}              & 0.8130  & 0.8060  & \textbf{0.8160} & --  \\
Li \etal \cite{li2018a2}                    & 0.8019 & 0.7961 & 0.7902 & 0.6633  \\\hline\hline
Baseline\_N                                 & \textbf{0.8237} & \textbf{0.8299} & 0.8079 & 0.6379  \\
Baseline\_C                                 & 0.7843 & 0.7599 & 0.7636 & 0.6647  \\\hline
\end{tabular}
\end{table}

The special properties of image cropping make it a challenging task to train an effective and efficient cropping model. On one hand, since the annotation of image cropping (which requires good knowledge and experience in photography) is very expensive \cite{chen2017quantitative}, existing cropping databases \cite{yan2013learning,fang2014automatic,chen2017quantitative} provide only one or several annotated crops for about 1,000 source images. On the other hand, the searching space of image cropping is very huge, with millions of candidate crops for each image. Clearly, the amount of annotated data in current databases is insufficient to train a robust cropping model.

In this work, we reconsider the problem of image cropping and propose a new approach, namely grid anchor based image cropping, to address this challenging task in a reliable and efficient manner. Our contributions are threefold.
\begin{enumerate}
\vspace{-6pt}
\item[1).] We propose a grid anchor based formulation for image cropping by considering the special properties and requirements of this problem. Our formulation reduces the number of candidate crops from millions to less than one hundred, providing a very efficient solution for image cropping.

\vspace{-6pt}
\item[2).] Based on our formulation, we construct a new image cropping database with exhaustive annotations for each source image. With 106,860 annotated candidate crops, our database provides a good platform to learn robust image cropping models. More reliable metrics are also defined to evaluate the performance of learned cropping models.

\vspace{-6pt}
\item[3).] We design an efficient and effective module for image cropping under the convolutional neural network (CNN) architecture. The learned cropping model runs at a speed of 125 FPS and obtains promising performance under various requirements.

\end{enumerate}

\section{Related work}

The existing image cropping methods can be divided into three categories according to their major drives.

\textbf{Attention-driven methods.} Earlier methods are mostly attention-driven, aiming to identify the major subject or the most informative region of an image. Most of them \cite{chen2003visual,suh2003automatic,stentiford2007attention,marchesotti2009framework} resort to a saliency detection algorithm (e.g. \cite{itti1998model}) to get an attention map of an image, and search a cropping window with the highest attention value. Some methods also employ face detection \cite{zhang2005auto} or gaze interaction \cite{santella2006gaze} to find the important region of an image.

\textbf{Aesthetic-driven methods.} The aesthetic-driven methods improve the attention-based methods by emphasizing the overall aesthetic quality of images. These methods \cite{zhang2005auto,nishiyama2009sensation,cheng2010learning,liu2010optimizing,yan2013learning,zhang2013probabilistic,fang2014automatic,zhang2014weakly} usually design a set of hand-crafted features to characterize the image aesthetic properties or composition rules. Some methods  further deign quality measures \cite{zhang2005auto,liu2010optimizing} to evaluate the quality of candidate crops, while some resort to training an aesthetic discriminator such as SVM \cite{nishiyama2009sensation,cheng2010learning}. The release of two cropping databases \cite{yan2013learning,fang2014automatic} facilitates the training of discriminative cropping models. However, the handcrafted features are not strong enough to accurately predict image aesthetics \cite{deng2017image}.

\textbf{Data-driven methods.} Most recent methods are data-driven, which train an end-to-end CNN model for image cropping. However, limited by the insufficient number of annotated training samples, many methods in this category \cite{chen2017quantitative,wang2017deep,wang2018deep,deng2017image,deng2017aesthetic,guo2017automatic,li2018a2} adopt a general aesthetic classifier trained from image aesthetic databases such as AVA \cite{murray2012ava} and CUHKPQ \cite{luo2011content} to help cropping. However, a general aesthetic classifier trained on full images may not be able to reliably evaluate the crops within one image \cite{chen2017learning,wei2018good}. An alternative strategy is to use pairwise learning to construct more training data \cite{chen2017learning,wei2018good} . But annotation of ranking pairs is also very expensive because of the subjective nature of image cropping. Recently, Wei \etal \cite{wei2018good} constructed a large scale comparative photo composition (CPC) database using an efficient two-stage annotation protocol, which provides a good training set for pairwise learning. Unfortunately, pairwise learning cannot provide adequate evaluation metrics for image cropping.


\section{Grid anchor based image cropping}

As illustrated in Fig. \ref{figure:problems}, image cropping has a high degree of freedom. There is not a unique optimal crop for a given image. We consider two practical requirements of a good image cropping system. Firstly, a reliable cropping system should be able to return acceptable results for different settings (e.g., aspect ratio and resolution) rather than one single output. Secondly, the cropping system should be lightweight and efficient to run on resource limited devices. With these considerations, we propose a grid anchor based formulation for practical image cropping, and construct a new benchmark under this formulation.

\subsection{Grid anchor based formulation}

\begin{figure}[t]
\centering
\subfigure{
\begin{minipage}[b]{1.0\linewidth}
\centering
\includegraphics[width=0.9\textwidth]{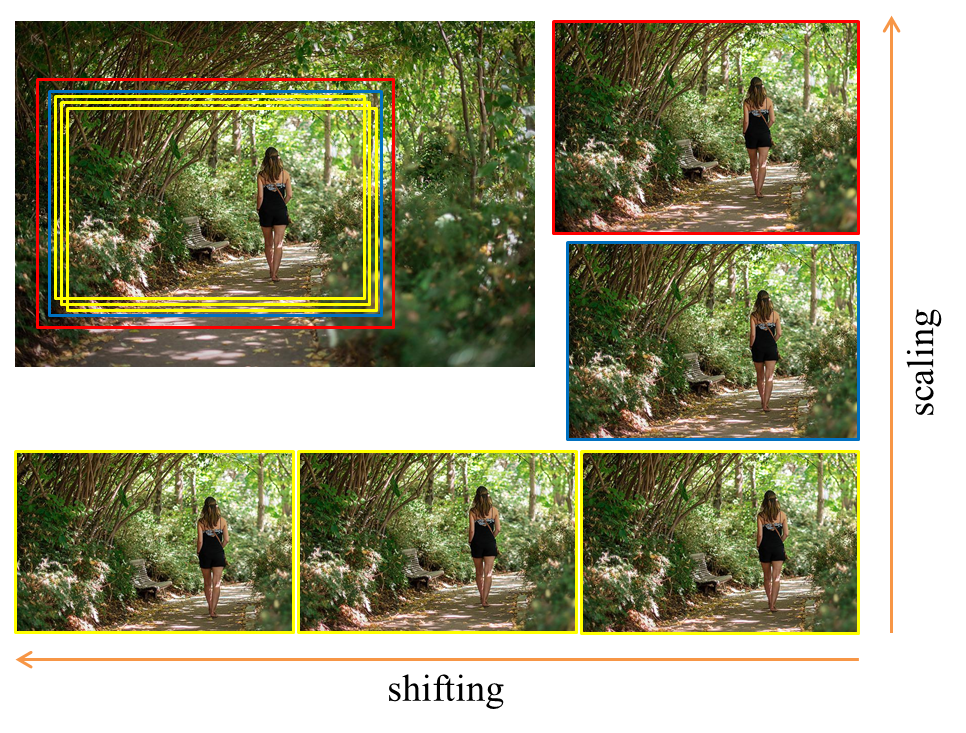}
\end{minipage}}%
\caption{The local redundancy of image cropping. Small local changes (e.g., shifting and/or scaling) on the cropping window of an acceptable crop (the bottom-right one) are very likely to output acceptable crops too.}
\label{figure:localInsensitivity}
\end{figure}

Given an image with resolution $H \times W$, a candidate crop can be defined using its top-left corner $(x_1,y_1)$ and bottom-right corner $(x_2,y_2)$, where $1\leq x_1<x_2\leq H$ and $1\leq y_1<y_2\leq W$. It is easy to calculate that the number of candidate crops is $\frac{H(H-1)W(W-1)}{4}$, which is a huge number even for an image of size $100\times100$. Fortunately, by exploiting the following properties and requirements of image cropping, the searching space can be significantly reduced, making automatic image cropping a tractable problem.

\begin{figure}[t]
\centering
\subfigure{
\begin{minipage}[b]{1.0\linewidth}
\centering
\includegraphics[width=0.9\textwidth]{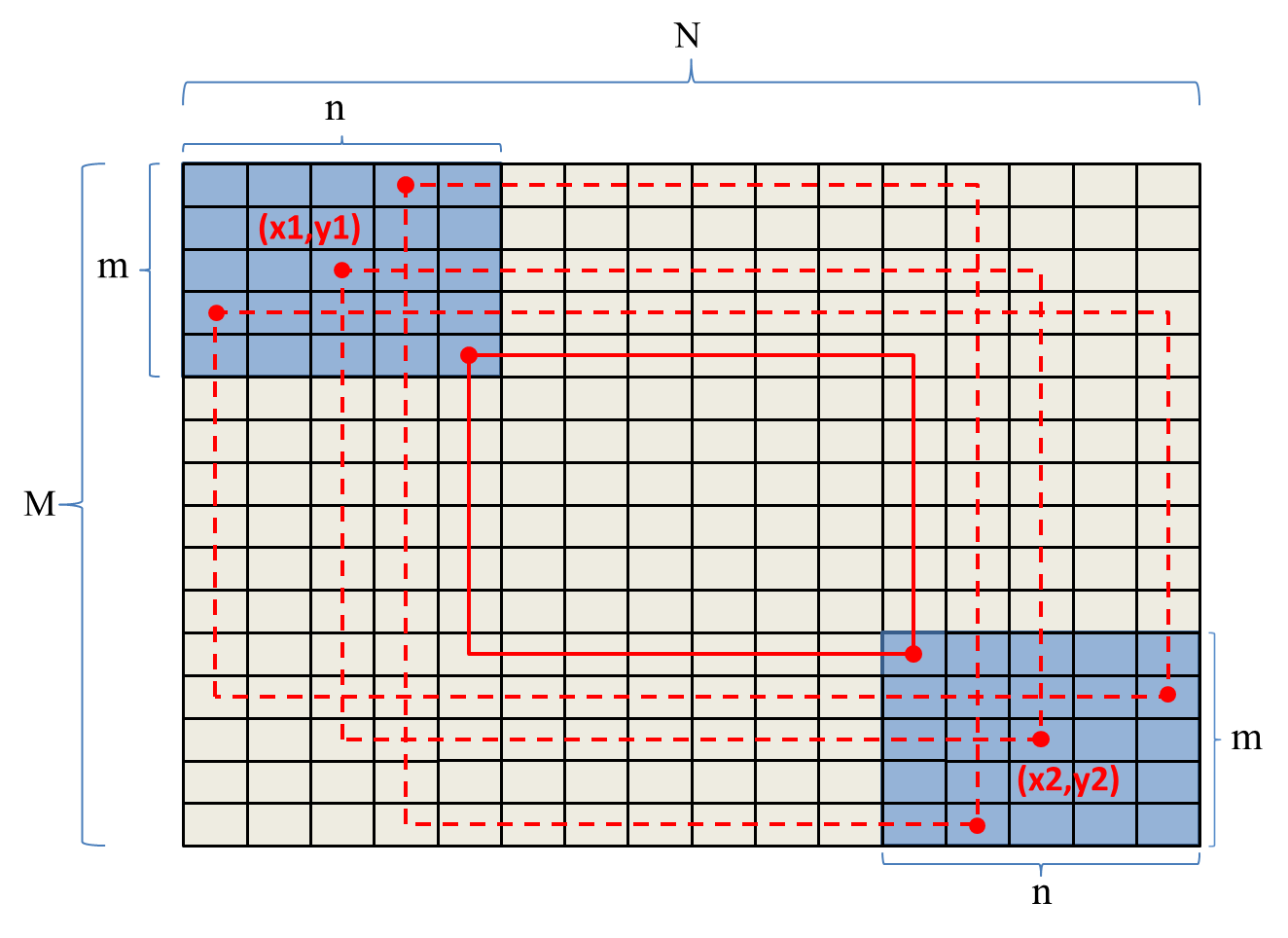}
\end{minipage}}%
\caption{Illustration of the grid anchor based formulation of image cropping. $M$ and $N$ are the numbers of bins for grid partition, while $m$ and $n$ define the adopted range of anchors for content preservation.}
\label{figure:simplification}
\end{figure}

\textit{Local redundancy:} Image cropping is naturally a problem with local redundancy. As illustrated in Fig. \ref{figure:localInsensitivity}, a set of similar and acceptable crops can be obtained in the neighborhood of a good crop by shifting and/or scaling the cropping widow. Intuitively, we can remove the redundant candidate crops by defining crops on image grid anchors rather than dense pixels. The proposed grid anchor based formulation is illustrated in Fig. \ref{figure:simplification}. We construct an image grid with $M \times N$ bins on the original image, and define the corners $(x_1,y_1)$ and $(x_2,y_2)$ of one crop on the grid centers, which serve as the anchors to generate a representative crop in the neighborhood. Such a formulation largely reduces the number of candidate crops from $\frac{H(H-1)W(W-1)}{4}$  to $\frac{M(M-1)N(N-1)}{4}$, which can be several orders smaller.

\textit{Content preservation:} Generally, a good crop should preserve the major content of the source image \cite{fang2014automatic}. Therefore, the cropping window should not be too small in order to avoid discarding too much the image content. To this end, we constrain the anchor points $(x_1,y_1)$ and $(x_2,y_2)$ of a crop into two regions with $m \times n$ bins on the top-left and bottom-right corners of the source image, respectively, as illustrated in Fig. \ref{figure:simplification}. This further reduces the number of crops from $\frac{M(M-1)N(N-1)}{4}$ to $m^2n^2$.

The smallest possible crop (highlighted in red solid lines in Fig. \ref{figure:simplification}) generated by the proposed scheme covers about $\frac{(M-2m+1)(N-2n+1)}{MN}$ grids of the source image, which may still be too small to preserve enough image content. We thus further constrain the area of potential crops to be no smaller than a certain proportion of the whole area of source image:
\begin{equation}\label{equ:area constraint}
S_{crop} \geq \lambda S_{Image},
\end{equation}
where $S_{crop}$ and $S_{Image}$ represent the areas of crop and original image, and $\lambda \in [\frac{(M-2m+1)(N-2n+1)}{MN},1)$.

\textit{Aspect ratio:} Because of the standard resolution of imaging sensors and displays, most people have been accustomed to the popular aspect ratios such as 16:9, 4:3 and 1:1. Candidate crops which have very different aspect ratios may be inconvenient to display and can make people feel uncomfortable. We thus require the aspect ratio of acceptable candidate crops satisfy the following condition:
\begin{equation}\label{equ:aspect ratio constraint}
\alpha_1 \leq \frac{W_{crop}}{H_{crop}} \leq \alpha_2,
\end{equation}
where $W_{crop}$ and ${H_{crop}}$ are the width and height of a crop. $\alpha_1$ and $\alpha_2$ define the range of aspect ratio and we set them to 0.5 and 2 to cover most common aspect ratios.

With Eq. \ref{equ:area constraint} and Eq. \ref{equ:aspect ratio constraint}, the final number of candidate crops in each image is less than $m^2n^2$.

\subsection{Grid anchor based cropping database}

Our proposed grid anchor based formulation reduces the number of candidate crops from $\frac{H(H-1)W(W-1)}{4}$ to less than $m^2n^2$. This enables us to annotate all the candidate crops for each image. To make the annotation cost as low as possible, we first made a small scale subjective study to find the smallest \{$M,N,m,n$\} that ensure at least 3 acceptable crops for each image. We collected 100 natural images and invited five volunteers to participate in this study. We set $M=N$ $\in \{16,14,12,10\}$ and $m=n$ $\in \{5,4,3\}$ to reduce possible combinations. $\lambda$ in Eq.\ref{equ:area constraint} was set to 0.5. After the tests, we found that $M=N=12$ and $m=n=4$ can lead to a good balance between cropping quality and annotation cost. Finally, the number of candidate crops is successfully reduced to no more than 90 for each image. Note that the setting of these parameters mainly aims to reduce annotation cost for training. In the testing stage, it is straightforward to use finer image grid to generate more candidate crops.

With the above settings, we constructed a Grid Anchor based Image Cropping Database (GAICD). We first crawled $\sim$50,000 images from the Flickr website. Considering that many images uploaded to Flickr already have good composition, we manually selected 1,000 images whose composition can be obviously improved, as well as 236 images with proper composition to ensure the generality of the GAICD. The selected images cover a variety of scenes and lighting conditions. For each image, our annotation toolbox (please refer to the supplementary file for details) automatically generates all the candidate crops in ordered aspect ratio. There are 106,860 candidate crops of the 1,236 images in total. The annotators were required to rate the candidates at five scores (from 1 to 5) which represent ``bad," ``poor," ``fair," ``good," and ``excellent".

A total of 19 annotators passed our test on photography composition and participated into the annotation. They are either experienced photographers from photography communities or students from the art department of two universities. Each crop was annotated by seven different subjects. The mean opinion score (MOS) was calculated for each candidate crop as its groundtruth quality score. We found that for 94.25\% candidate crops in our database, the standard deviations of their rating scores are smaller than 1, which confirms the annotation consistency under our grid anchor based formulation. More statistical analyses of our GAICD are presented in the supplementary file. Fig. \ref{figure:database} shows one source image and several of its annotated crops (with MOS scores) in the GAICD.

\begin{figure}[t]
\centering
\subfigure{
\begin{minipage}[b]{1.0\linewidth}
\centering
\includegraphics[width=1.0\textwidth]{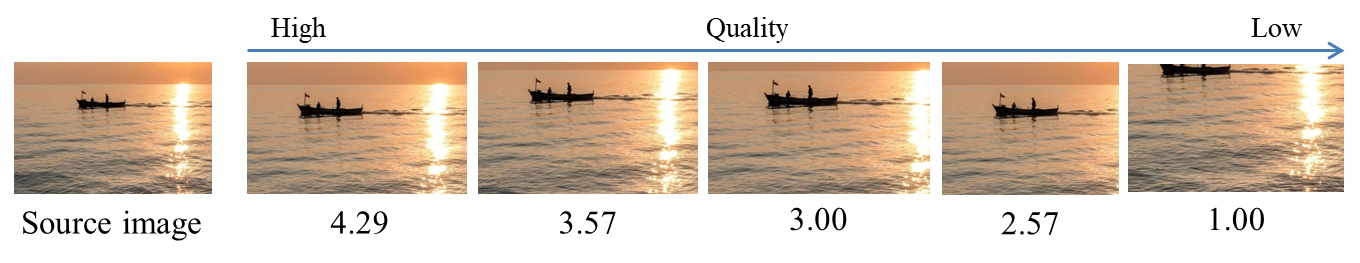}
\end{minipage}}%
\caption{One example source image and several of its annotated crops in our GAICD. The MOS is marked under each crop.}
\label{figure:database}
\end{figure}

\subsection{Evaluation metrics}\label{Evaluation metrics}

The dense annotations of our GAICD enable us to define more reliable metrics to evaluate cropping performance than IoU or BDE used in previous databases \cite{yan2013learning,fang2014automatic,chen2017quantitative}. We define two metrics on GAICD. The first one is average Spearman's rank-order correlation coefficient (SRCC). The SRCC has been widely used to evaluate the rank correlation between the MOS and model's predictions in image quality and aesthetic assessment \cite{kong2016photo, ma2017waterloo}.  Denote by $\mathbf{g}_i$ the vector of MOS of all crops for image $i$, and by $\mathbf{p}_i$ the predicted scores of these crops by a model. The SRCC is defined as:
\begin{equation}\label{equ:srcc}
SRCC(\mathbf{g}_i,\mathbf{p}_i) = cov(\mathbf{r}_{\mathbf{g}_i},\mathbf{r}_{\mathbf{p}_i}) / (std(\mathbf{r}_{\mathbf{g}_i})std(\mathbf{r}_{\mathbf{p}_i})),
\end{equation}
where $\mathbf{r}_{\mathbf{g}_i}$ and $\mathbf{r}_{\mathbf{p}_i}$ record the ranking order of scores in $\mathbf{g}_i$ and $\mathbf{p}_i$, and $cov(\cdot)$ and $std(\cdot)$ are the operators of covariance and standard deviation. The average SRCC is defined as:
\begin{equation}\label{equ:avg srcc}
\overline{SRCC} = \frac{1}{T}\sum\nolimits_{i=1}^{T} SRCC(\mathbf{g}_i,\mathbf{p}_i),
\end{equation}
where $T$ is the number of testing images.

\begin{figure*}[t]
\centering
\subfigure{
\begin{minipage}[b]{1.0\linewidth}
\centering
\includegraphics[width=0.9\textwidth]{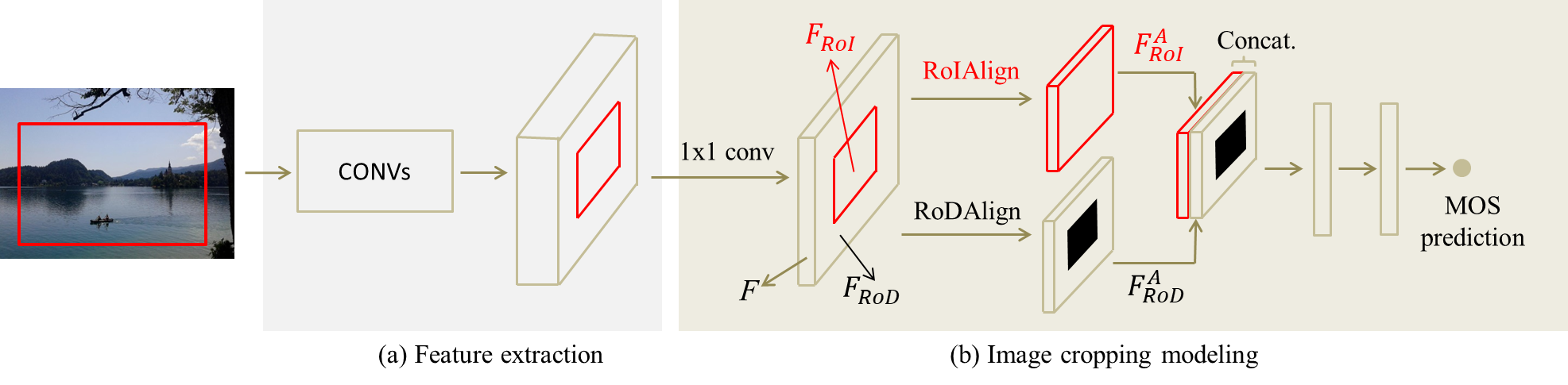}
\end{minipage}}%
\caption{The proposed CNN architecture for image cropping model learning.}
\label{figure:CNN}
\end{figure*}

Considering the fact that users may care more about whether the returned crops are acceptable or not than the accurate ranking order of all crops, we define a new metric, which we call ``return $K$ of top-$N$ accuracy" ($Acc_{K/N}$), for practical cropping applications. Denote by $S_{i}(N)$ the set of crops whose MOS rank the top-$N$ for image $i$, and denote by $\{c_{i1},c_{i2},...,c_{iK}\}$ the set of $K$ best crops returned by a cropping model. The $Acc_{K/N}$ aims to check how many of the $K$ returned crops fall into set $S_{i}(N)$:
\begin{equation}\label{equ:topN accuracy}
Acc_{K/N} = \frac{1}{TK}\sum\nolimits_{i=1}^{T}\sum\nolimits_{j=1}^{K} True(c_{ij} \in S_{i}(N)),
\end{equation}
where $True(*) = 1$ if * is true, otherwise $True(*) = 0$. In our experiments, we set $N$ to either 5 or 10, and evaluate $K$ = 1, 2, 3, 4 for both $N$ = 5 and $N$ = 10. We further average $Acc_{K/N}$ over $K$ for each $N$, leading to two average accuracy metrics:
\begin{equation}\label{equ:avg accuracy}
\overline{Acc_{N}} = \frac{1}{4}\sum\nolimits_{K=1}^{4} Acc_{K/N}.
\end{equation}

\section{Cropping model learning}

Limited by insufficient training data, most previous cropping methods focused on how to leverage additional aesthetic databases \cite{wang2017deep,chen2017learning,deng2017aesthetic} or how to construct more training pairs \cite{chen2017quantitative,wei2018good}, paying limited attention to how to design a suitable network for image cropping itself. They usually adopt the standard CNN architecture widely used in object detection. Our GAICD provides a better platform with much more annotated samples for model training. By considering the special properties of image cropping, we design an effective and lightweight module for cropping model learning. The overall architecture is shown in Fig. \ref{figure:CNN}, which consists of one general feature extraction module and one image cropping module.

\textbf{Feature extraction:} As in many previous works \cite{wang2017deep,deng2017image,chen2017quantitative,chen2017learning,deng2017aesthetic,guo2017automatic,li2018a2,wei2018good}, we truncate one pre-trained CNN model (e.g., VGG16 \cite{simonyan2014very} or ResNet50 \cite{he2016deep}) as the feature extraction module. The spatial arrangement of context and objects in an image plays a key role in image composition. For example, the ``rule of thirds", which is the most commonly used composition rule, suggests to place important compositional elements at certain locations of an image \cite{wiki:rule-of-thirds}. Therefore, the feature extraction module needs to preserve sufficient spatial resolution for evaluating image composition in the following cropping module. Truncating at shallower layers can preserve higher spatial resolution but the output feature map may not have enough receptive field to describe large objects in images. We conducted extensive experiments to decide the most cost-effective layer to truncate two standard CNN models for image cropping. More details can be found in Sec. \ref{Feature extraction module}.

\textbf{Modeling both the RoI and RoD:} One significant difference between image cropping and object detection is that object detection only focuses on the region of interest (RoI), while cropping also needs to consider the discarded information (hereafter we call it region of discard (RoD)). On one hand, removing distracting information can significantly improve the composition. On the other hand, cropping out important region can dramatically change or even destroy an image. Taking the second last crop in Fig. \ref{figure:database} as an example, although it may have acceptable composition but its visual quality is much lower than the source image because the beautiful sunset glow is cropped out. The discarded information is unavailable to the cropping model if only the RoI is considered, while modeling the RoD can effectively solve this problem.

Referring to Fig. \ref{figure:CNN}, let $F$ denote the whole feature map output by the feature extraction module, and the feature maps in RoI and RoD are denoted by $F_{RoI}$ and $F_{RoD}$, respectively. We first employ the RoIAlign \cite{he2017mask} to transform $F_{RoI}$ into $F_{RoI}^{A}$ which has fixed spatial resolution $s \times s$. The $F_{RoD}$ is constructed by removing $F_{RoI}$ from $F$, namely, setting the values of $F_{RoI}$ to zeros in $F$. Then the RoDAlign (using the same bilinear interpolation as RoIAlign) is performed on $F_{RoD}$, leading to $F_{RoD}^{A}$ which has the same spatial resolution as $F_{RoI}^{A}$. $F_{RoI}^{A}$ and $F_{RoD}^{A}$ are concatenated along the channel dimension as one aligned feature map which contains the information in both RoI and RoD. The combined feature map is fed into two fully connected layers for final MOS prediction.

\textbf{Reducing the channel dimension:} Another difference between image cropping and object detection is that the former does not need to accurately recognize the category of different objects, which allows us to significantly reduce the channel dimension of the feature map. In practice, we find that the channel dimension of the feature map (output by the VGG16 model) can be reduced from 512 to 8 using $1 \times 1$ convolution without sacrificing much the performance. The low channel dimension makes our image cropping module very efficient and lightweight. More details can be found in Sec. \ref{Feature extraction module}.

\textbf{Loss function:} Denote by $e_{ij}=g_{ij}-p_{ij}$, where $g_{ij}$ and $p_{ij}$ are the groundtruth MOS and predicted score of the $j$-th crop for image $i$. The Huber loss \cite{huber1964robust} is employed as the loss function to learn our cropping model because of its robustness to outliers:
\begin{equation}\small \label{equ:Huber}
\mathcal{L}_{ij}=
\left\{
\begin{aligned}
&\frac{1}{2}e_{ij}^2, \textrm{when}\; |e_{ij}|\leq\delta,\\
&\delta|e_{ij}|-\frac{1}{2}\delta^2, \textrm{otherwise},
\end{aligned}
\right.
\end{equation}
where $\delta$ is fixed at 1 throughout our experiments.

\section{Experiments}

\subsection{Implementation details}

We randomly selected 200 images from our GAICD as the testing set and used the remaining 1,036 images (containing 89,519 annotated crops in total) for training and validation. In the training stage, our model takes one image and 64 randomly selected crops of it as a batch to input. In the testing stage, the trained model evaluates all the generated crops of one image and outputs a predicted MOS for each crop. To improve the training and testing efficiency, the short side of input images is resized to 256. The standard ADAM optimizer with the default parameters was employed to train our model for 40 epoches. Learning rate was fixed at $1e^{-4}$ throughout our experiments. We randomly adjusted the contrast and saturation of the source images for data augmentation in the training stage. The MOS were normalized by removing the mean and dividing by the standard deviation across the training set.

\subsection{Ablation study of our cropping model}

\subsubsection{Feature extraction module}\label{Feature extraction module}

\begin{table}[t]
\footnotesize
\centering
\caption{Image cropping performance by using different feature extraction modules. The truncating layer (tlayer), stride (str), receptive field (rf) and parameter size (par (Mbit)) of the feature extraction module are shown for each case.}
\label{table:model}
\begin{tabular}{p{8mm}|p{5mm}ccc|c|c|c}
\hline
& & & & & & &\\[-1em]
model &tlayer &str & rf & par & $\overline{SRCC}$ & $\overline{Acc_{5}}$ & $\overline{Acc_{10}}$  \\\hline
\multirow{5}{*}{vgg16} &c4\_1&8&60 &11.1& 0.695 & 40.1 & 58.3 \\
&c4\_3&8&92 &29.1    & 0.715 & 42.5 & 61.8 \\
&c5\_1&16&132 &38.1  & 0.735 & 46.6 & 65.5  \\
&c5\_3&16& 192 &56.1 & \textbf{0.737} & \textbf{47.0} & \textbf{65.6} \\
&pool5&32&212 &56.1  & 0.702 & 43.6 & 61.9 \\\hline
\multirow{5}{*}{resnet50} &c3\_2&8&67& $\ \ $3.4& 0.620 & 33.1 & 50.8 \\
&c3\_4&8&99& $\ \ $5.6& 0.647 & 35.1 & 52.9 \\
&c4\_3&16&195 &19.9   & 0.709 & 41.8 & 60.8 \\
&c4\_6&16&291 &32.7   & \textbf{0.712} & \textbf{42.1} & \textbf{61.2}  \\
&c5\_1&32&355 &55.8   & 0.692 & 40.6 & 58.3 \\\hline
\end{tabular}
\end{table}

\begin{table}[t]
\footnotesize
\centering
\caption{Ablation experiments on the RoI and RoD.}
\label{table:roirod}
\begin{tabular}{c|c|c|c}
\hline
& & & \\[-1em]
module      & $\overline{SRCC}$ & $\overline{Acc_{5}}$ &$\overline{Acc_{10}}$  \\\hline
RoD         & 0.597 &  29.8 &  43.4  \\
RoI         & 0.706 &  44.8 &  62.9  \\
RoI+RoD     & \textbf{0.735} & \textbf{46.6} &  \textbf{65.5}   \\\hline
\end{tabular}
\end{table}

\begin{table}[t]
\footnotesize
\centering
\caption{Image cropping performance by using different spatial resolution ($s \times s$) and channel dimension (cdim). The number of filters (nfilter) is fixed as 512 in the FC layers.  The VGG16 model (truncated at conv5\_1) is employed as the feature extraction module for all cases. The parameter size (par (Mbit)) of the image cropping module (including two FC layers with $s \times s \times (2*cdim) \times 512$ and $1\times 1\times 512 \times 512$ kernels) is reported for each case.}
\label{table:warpingandFC}
\begin{tabular}{c|c|c|c|c|c|c}
\hline
& & & & & & \\[-1em]
$s \times s$ & cdim & nfilter & par & $\overline{SRCC}$ & $\overline{Acc_{5}}$ & $\overline{Acc_{10}}$  \\\hline
3$\times$3   & 8  & 512 & 1.28  & 0.689 & 42.4 & 58.9    \\
5$\times$5   & 8  & 512 & 1.78  & 0.711 & 44.6 & 61.5    \\
7$\times$7   & 8  & 512 & 2.53  & 0.725 & 45.4 & 63.1 \\
9$\times$9   & 8  & 512 & 3.53  & 0.735 & 46.6 & 65.5  \\
11$\times$11 & 8  & 512 & 4.78  & \textbf{0.736} & \textbf{46.8} & \textbf{65.6}  \\\hline
9$\times$9   & 32 & 512 & 11.13 & 0.733 & 46.4 & 65.3\\
9$\times$9   & 16 & 512 & 6.06  & \textbf{0.736} & \textbf{46.8} & \textbf{65.8}  \\
9$\times$9   & 8  & 512 & 3.53  & 0.735 & 46.6 & 65.5    \\
9$\times$9   & 4  & 512 & 2.27  & 0.731 & 45.9 & 65.1 \\
9$\times$9   & 2  & 512 & 1.63  & 0.719 & 45.1 & 64.1  \\
9$\times$9   & 1  & 512 & 1.32  & 0.706 & 43.8 & 62.6   \\\hline
\end{tabular}
\end{table}

We first conduct a set of experiments to determine the appropriate feature extraction module on two pre-trained models (VGG16 \cite{simonyan2014very} and ResNet50 \cite{he2016deep}). For each model, we truncated at five different layers, which cover various strides and receptive fields, and evaluated their effects on cropping performance. The image cropping module (including both the RoI and RoD) was fixed for all cases. The truncating layer, stride, receptive field, parameter size and cropping performance for each module are reported in Table \ref{table:model}. To save space, we do not report each single accuracy index in the ablation study.

We can make three observations from Table \ref{table:model}. First, for both the VGG16 and ResNet50 models, a too small receptive field in the feature extraction module will lead to unsatisfied performance. Increasing the receptive field can significantly improve the cropping accuracy at the cost of deeper architecture and more parameters. The performance plateaus when the receptive field is increased to more than half of the image size. It is worth noting that the above observations on stride and receptive field are based on certain input image size (short side equals to 256 in our experiments), which may provide good reference for other input size. Second, a too large stride (e.g., 32) deteriorates the performance, either. This is because downsampling too much the feature map will lose important spatial information for image cropping. Specifically, for the input image of resolution $256\times256$, downsampling with stride 32 will result in feature maps of size $8\times8$, and consequently the feature map of a candidate crop may only have a spatial resolution of $4\times4$, which is insufficient to generate accurate crops. Finally, the VGG16 models generally outperforms the ResNet50 models. This may be because the ResNet50 models can be overfitted on our database. We thus choose the VGG16 model (truncated at conv5\_1 layer) as the feature extraction module in the following experiments.

\vspace{-8pt}
\subsubsection{Image cropping module}

We then evaluate the proposed image cropping module, including the effects of parameter size, RoI and RoD.

\textbf{Parameter size:} There are two key parameters in the image cropping module: spatial resolution ($s \times s$) of the aligned feature map and channel dimension (cdim) after dimension reduction. Table \ref{table:warpingandFC} reports the cropping performance of using different $s \times s$ and cdim. The number of filters was fixed at 512 for the FC layers. We first found that a smaller $s$ (e.g. 3 or 5) would result in obviously worse performance. This again proves the importance of sufficient spatial information for image cropping. $s = 9$ seems to be an appropriate choice since further increasing the value does not bring obvious improvements. The channel dimension of feature maps can be significantly reduced for the problem of image cropping. As can be seen from Table \ref{table:warpingandFC}, the performance is still reasonable even if we reduce the channel dimension to 1 (note that VGG16 output 512 channels of feature maps). The low channel dimension makes the proposed image cropping module efficient and lightweight. In the following experiments, we chose 8 as the reduced channel dimension which has a good trade-off between cost and efficacy. Under this setting, the whole image cropping module has only 3.53 Mbits parameters.

\textbf{RoI and RoD:} We make an ablation study on the role of RoI and RoD. The results of using only RoI, only RoD and both of them are reported in Table \ref{table:roirod}. As can be seen, modeling only the RoD results in very poor accuracy, modeling only the RoI performs much better, while modeling simultaneously the RoI and RoD achieves the best cropping accuracy in all cases. This corroborates our analysis that image cropping needs to consider both the RoI and RoD.

\begin{figure}[t]
\centering
\subfigure{
\begin{minipage}[b]{1.0\linewidth}
\centering
\includegraphics[width=1.0\textwidth]{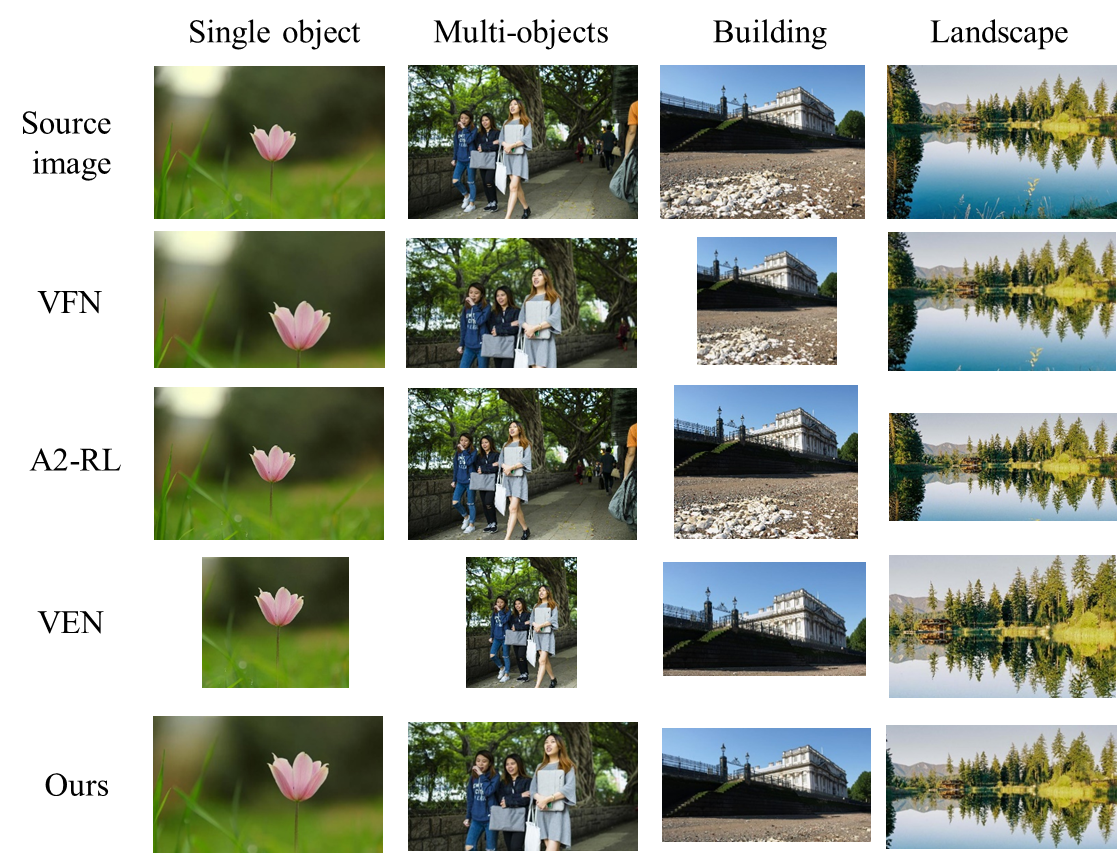}
\end{minipage}}%
\caption{Qualitative comparison of returned top-1 crop by different methods.}
\label{figure:qua comp top1}
\end{figure}

\begin{figure*}[t]
\centering
\subfigure{
\begin{minipage}[b]{1.0\linewidth}
\centering
\includegraphics[width=1.0\textwidth]{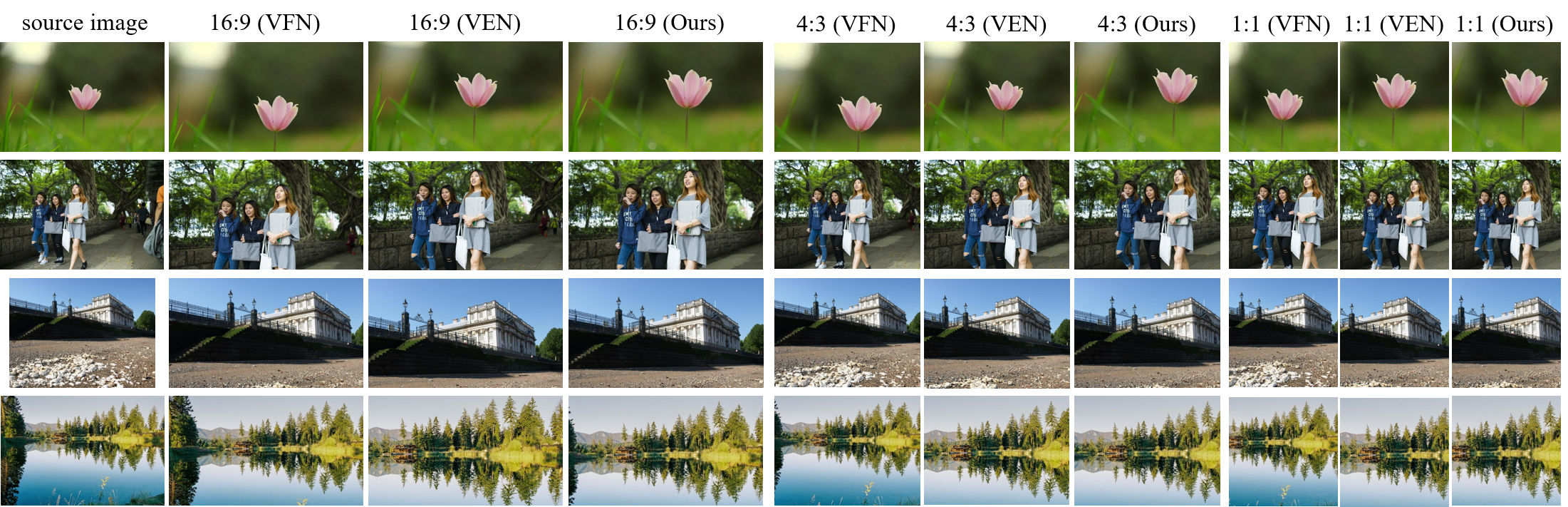}
\end{minipage}}%
\caption{Qualitative comparison of returning crops with different aspect ratios by different methods.}
\label{figure:qua comp fixed}
\end{figure*}

\begin{table*}[t]
\footnotesize
\centering
\caption{Quantitative comparison between different methods on the GAICD. ``--" means that result is not available.}
\label{table:others}
\begin{tabular}{c|c|cccc|c|cccc|c|c}
\hline
& & & & & & & & & & & &\\[-1em]
Method                     & $\overline{SRCC}$ & $Acc_{1/5}$ & $Acc_{2/5}$ & $Acc_{3/5}$ & $Acc_{4/5}$ & $\overline{Acc_{5}}$ & $Acc_{1/10}$ & $Acc_{2/10}$ & $Acc_{3/10}$ & $Acc_{4/10}$ &$\overline{Acc_{10}}$ & FPS \\\hline
Baseline\_L                & --   & 24.5 & --   & --   & --   & --   & 41.0 & --   & --   & --   & --  & -- \\
A2-RL \cite{li2018a2}      & --   & 23.0 & --   & --   & --   & --   & 38.5 & --   & --   & --   & --  & 4 \\
VPN\cite{wei2018good}      & --   & 40.0 & --   & --   & --   & --   & 49.5 & --   & --   & --   & --  & 75 \\
VFN\cite{chen2017learning} & 0.450 & 27.0 & 30.0 & 26.0 & 17.5 & 25.1 & 39.0 & 40.5 & 39.0 & 31.5 & 37.5 & 0.5\\
VEN\cite{wei2018good}      & 0.621 & 40.5 & 37.5 & 38.5 & 36.5 & 38.1 & 54.0 & 51.5 & 50.5 & 47.0 & 50.8 & 0.2\\
GAIC (ours)                & \textbf{0.735} & \textbf{53.5} & \textbf{47.0} & \textbf{44.5} & \textbf{41.5} & \textbf{46.6} & \textbf{71.5} & \textbf{66.0} & \textbf{66.5} & \textbf{58.0} & \textbf{65.5} & \textbf{125}  \\\hline
\end{tabular}
\end{table*}

\subsection{Comparison to other methods}

As discussed in the introduction section, the limitations of existing image cropping databases and evaluation metrics make the learning and evaluation of reliable cropping models difficult. Nonetheless, we still evaluated our model on the previous databases \cite{yan2013learning,chen2017quantitative}, and the results can be found in the \textbf{supplementary} file. Here we report the experimental results on the proposed GAICD.

\vspace{-10pt}
\subsubsection{Comparison methods}

Though a number of image cropping methods have been developed \cite{wang2017deep,deng2017image,chen2017quantitative,chen2017learning,deng2017aesthetic,guo2017automatic,li2018a2,wei2018good}, many of them do not release the source code or executable program. We thus compare our method, namely Grid Anchor based Image Cropping (GAIC), with the following baseline and recently developed state-of-the-art methods whose source codes are available.

\textbf{Baseline\_L:} The baseline\_L does not need any training. It simply outputs the largest crop among all eligible candidates. The result is similar to the ``baseline\_N" mentioned in Table \ref{table:Performance comparison}, i.e., the source image without cropping.

\textbf{VFN \cite{chen2017learning}:} The View Finding Network (VFN) is trained in a pair-wise ranking manner using professional photographs crawled from the Flickr. High-quality photos were first manually selected, and a set of crops were then generated from each image. The ranking pairs were constructed by always assuming that the source image has better quality than the generated crops.

\textbf{VEN and VPN \cite{wei2018good}:} Compared with VFN, the View Evaluation Network (VEN) employs more reliable ranking pairs to train the model. Specifically, the authors annotated more than 1 million ranking pairs using a two-stage annotation strategy. A more efficient View Proposal Network (VPN) was proposed in the same work, and it was trained using the predictions of VEN. The VPN is based on the detection model SSD \cite{liu2016ssd}, and it outputs a prediction vector for 895 predefined boxes.

\textbf{A2-RL \cite{li2018a2}:} The A2RL is trained in an iterative optimization manner. The model adjusts the cropping window and calculates a reward (based on predicted aesthetic score) for each step. The iteration stops when the accumulated reward satisfies some termination criteria.

\vspace{-8pt}
\subsubsection{Qualitative comparison}

To demonstrate the advantages of our cropping method over previous ones, we first conduct qualitative comparison of different methods on four typical scenes: single object, multi-objects, building and landscape. Note that these images are out of any existing cropping databases.
In the first set of comparison, we compare all methods under the setting of returning only one best crop. Each model uses its default candidate crops generated by its source code except for VFN, which does not provide such code and uses the same candidates as our method. The results are shown in Fig. \ref{figure:qua comp top1}. We can make several interesting observations. Both VFN and A2-RL fail to robustly remove distracting elements in images. VFN cuts some important content, while A2-RL simply returns the source image in many cases. VEN and our GAIC model can stably output visually pleasing crops. The major differences lie in that VEN prefers more close-up crops while our GAIC tends to preserve as much useful information as possible.

A flexible cropping system should be able to output acceptable results under different requirements in practice, e.g., different aspect ratios. In Fig. \ref{figure:qua comp fixed}, we show the cropping results by the competing methods under three most commonly used aspect ratios: 16:9, 4:3 and 1:1. The A2-RL is not included because it does not support this test. Again, our model outputs the most visually pleasing crop in most cases. More results can be found in supplementary file.

\vspace{-8pt}
\subsubsection{Quantitative comparison}
\vspace{-4pt}

We then perform quantitative comparisons by using the metrics defined in Section \ref{Evaluation metrics}.  Among the competitors, VFN, VEN and our GAIC support predicting scores for all the candidate crops provided by our database, thus they can be quantitatively evaluated by all the defined evaluation metrics. VPN uses its own pre-defined cropping boxes which are different from our database, and Baseline\_L and A2-RL output only one single crop. Therefore, we can only calculate $Acc_{1/5}$ and $Acc_{1/10}$ for them. We approximate the output boxes by VPN and A2-RL to the nearest anchor box in our database when calculating the quantitative indexes.

The results of all competing methods are shown in Table \ref{table:others}. We can see that both A2-RL and VFN only obtain comparable performance to Baseline\_L. This is mainly because A2-RL is supervised by a general aesthetic classifier in training, and the ranking pairs used in VFN are not very reliable. By using more reliable ranking pairs, VEN obtains much better performance than VFN. VPN performs slightly worse than VEN as expected because it is supervised by the predictions of VEN. Our method outperforms VEN by a large margin, which owes to the richer cropping information leveraged by our annotation approach compared to the pair-wise ranking annotations used by VEN, as well as the more effective cropping module training of our model.

\vspace{-8pt}
\subsubsection{Running speed}
\vspace{-4pt}

A practical image cropping model should also have fast speed for real-time implementation. In the last column of Table \ref{table:others}, we compare the running speed in terms of frame-per-second (FPS) for all competing methods. All models are run on the same PC with i7-6800K CPU, 64G RAM and one GTX 1080Ti GPU. As can be seen, our GAIC model runs at 125 FPS, which is much faster than all the competitors. It is worth mentioning that both GAIC and VPN are based on VGG16 architecture, but GAIC has much less parameters than VPN ($40$ Mbits vs. $290$ Mbits). The other methods are much slower because A2-RL needs to iterate the cropping window while VFN and VEN need to individually process each crop.

\section{Conclusion}

We analyzed the limitations of existing formulation and databases on image cropping. Consequently, we proposed a more reliable and efficient formulation for practical image cropping, namely grid anchor based image cropping (GAIC). A new benchmark was constructed, which contains 1,236 source images and 106,860 annotated crops, as well as two types of reliable evaluation metrics. We further proposed a lightweight and effective cropping module under the CNN architecture. Our GAIC can robustly output visually pleasing crops under different aspect ratios and it runs at a speed of 125FPS, much faster than other methods.

{\footnotesize
\bibliographystyle{ieee}
\bibliography{egbib}

\begin{thebibliography}{10}\itemsep=-1pt

\bibitem{bhatt2015multifunctional}
N.~Bhatt and T.~Cherna.
\newblock Multifunctional environment for image cropping, Oct.~13 2015.
\newblock US Patent 9,158,455.

\bibitem{chedeau2017image}
C.~S.~B. Chedeau.
\newblock Image cropping according to points of interest, Mar.~28 2017.
\newblock US Patent 9,607,235.

\bibitem{chen2016automatic}
J.~Chen, G.~Bai, S.~Liang, and Z.~Li.
\newblock Automatic image cropping: A computational complexity study.
\newblock In {\em CVPR}, pages 507--515, 2016.

\bibitem{chen2003visual}
L.-Q. Chen, X.~Xie, X.~Fan, W.-Y. Ma, H.-J. Zhang, and H.-Q. Zhou.
\newblock A visual attention model for adapting images on small displays.
\newblock {\em Multimedia systems}, 9(4):353--364, 2003.

\bibitem{chen2017quantitative}
Y.-L. Chen, T.-W. Huang, K.-H. Chang, Y.-C. Tsai, H.-T. Chen, and B.-Y. Chen.
\newblock Quantitative analysis of automatic image cropping algorithms: A
  dataset and comparative study.
\newblock In {\em WACV}, pages 226--234, 2017.

\bibitem{chen2017learning}
Y.-L. Chen, J.~Klopp, M.~Sun, S.-Y. Chien, and K.-L. Ma.
\newblock Learning to compose with professional photographs on the web.
\newblock In {\em ACM Multimedia}, pages 37--45, 2017.

\bibitem{cheng2010learning}
B.~Cheng, B.~Ni, S.~Yan, and Q.~Tian.
\newblock Learning to photograph.
\newblock In {\em ACM Multimedia}, pages 291--300, 2010.

\bibitem{chor2006system}
A.~Chor, J.~Schwartz, P.~Hellyar, T.~Kasperkiewicz, and D.~Parlin.
\newblock System for automatic image cropping based on image saliency, Apr.~6
  2006.
\newblock US Patent App. 10/956,628.

\bibitem{ciocca2007self}
G.~Ciocca, C.~Cusano, F.~Gasparini, and R.~Schettini.
\newblock Self-adaptive image cropping for small displays.
\newblock {\em IEEE Transactions on Consumer Electronics}, 53(4), 2007.

\bibitem{deng2017aesthetic}
Y.~Deng, C.~C. Loy, and X.~Tang.
\newblock Aesthetic-driven image enhancement by adversarial learning.
\newblock {\em arXiv preprint arXiv:1707.05251}, 2017.

\bibitem{deng2017image}
Y.~Deng, C.~C. Loy, and X.~Tang.
\newblock Image aesthetic assessment: An experimental survey.
\newblock {\em IEEE Signal Processing Magazine}, 34(4):80--106, 2017.

\bibitem{downing2015automated}
E.~O. Downing, O.~M. Koenders, and B.~T. Grover.
\newblock Automated image cropping to include particular subjects, Apr.~28
  2015.
\newblock US Patent 9,020,298.

\bibitem{fang2014automatic}
C.~Fang, Z.~Lin, R.~Mech, and X.~Shen.
\newblock Automatic image cropping using visual composition, boundary
  simplicity and content preservation models.
\newblock In {\em ACM Multimedia}, pages 1105--1108, 2014.

\bibitem{freixenet2002yet}
J.~Freixenet, X.~Mu{\~n}oz, D.~Raba, J.~Mart{\'\i}, and X.~Cuf{\'\i}.
\newblock Yet another survey on image segmentation: Region and boundary
  information integration.
\newblock In {\em ECCV}, pages 408--422, 2002.

\bibitem{guo2017automatic}
G.~Guo, H.~Wang, C.~Shen, Y.~Yan, and H.-Y.~M. Liao.
\newblock Automatic image cropping for visual aesthetic enhancement using deep
  neural networks and cascaded regression.
\newblock {\em arXiv preprint arXiv:1712.09048}, 2017.

\bibitem{he2017mask}
K.~He, G.~Gkioxari, P.~Doll{\'a}r, and R.~Girshick.
\newblock Mask {R-CNN}.
\newblock In {\em ICCV}, pages 2980--2988. IEEE, 2017.

\bibitem{he2016deep}
K.~He, X.~Zhang, S.~Ren, and J.~Sun.
\newblock Deep residual learning for image recognition.
\newblock In {\em CVPR}, pages 770--778, 2016.

\bibitem{huber1964robust}
P.~J. Huber et~al.
\newblock Robust estimation of a location parameter.
\newblock {\em The annals of mathematical statistics}, 35(1):73--101, 1964.

\bibitem{itti1998model}
L.~Itti, C.~Koch, and E.~Niebur.
\newblock A model of saliency-based visual attention for rapid scene analysis.
\newblock {\em IEEE Transactions on pattern analysis and machine intelligence},
  20(11):1254--1259, 1998.

\bibitem{jogo2007image}
N.~Jogo.
\newblock Image cropping and synthesizing method, and imaging apparatus,
  Apr.~24 2007.
\newblock US Patent 7,209,149.

\bibitem{kong2016photo}
S.~Kong, X.~Shen, Z.~Lin, R.~Mech, and C.~Fowlkes.
\newblock Photo aesthetics ranking network with attributes and content
  adaptation.
\newblock In {\em ECCV}, pages 662--679. Springer, 2016.

\bibitem{li2018a2}
D.~Li, H.~Wu, J.~Zhang, and K.~Huang.
\newblock A2-{RL}: Aesthetics aware reinforcement learning for image cropping.
\newblock In {\em CVPR}, pages 8193--8201, 2018.

\bibitem{liu2010optimizing}
L.~Liu, R.~Chen, L.~Wolf, and D.~Cohen-Or.
\newblock Optimizing photo composition.
\newblock In {\em Computer Graphics Forum}, volume~29, pages 469--478, 2010.

\bibitem{liu2016ssd}
W.~Liu, D.~Anguelov, D.~Erhan, C.~Szegedy, S.~Reed, C.-Y. Fu, and A.~C. Berg.
\newblock {SSD}: Single shot multibox detector.
\newblock In {\em ECCV}, pages 21--37. Springer, 2016.

\bibitem{luo2011content}
W.~Luo, X.~Wang, and X.~Tang.
\newblock Content-based photo quality assessment.
\newblock In {\em ICCV}, pages 2206--2213, 2011.

\bibitem{ma2017waterloo}
K.~Ma, Z.~Duanmu, Q.~Wu, Z.~Wang, H.~Yong, H.~Li, and L.~Zhang.
\newblock Waterloo exploration database: New challenges for image quality
  assessment models.
\newblock {\em IEEE Transactions on Image Processing}, 26(2):1004--1016, 2017.

\bibitem{marchesotti2009framework}
L.~Marchesotti, C.~Cifarelli, and G.~Csurka.
\newblock A framework for visual saliency detection with applications to image
  thumbnailing.
\newblock In {\em ICCV}, pages 2232--2239, 2009.

\bibitem{murray2012ava}
N.~Murray, L.~Marchesotti, and F.~Perronnin.
\newblock {AVA}: A large-scale database for aesthetic visual analysis.
\newblock In {\em CVPR}, pages 2408--2415, 2012.

\bibitem{nishiyama2009sensation}
M.~Nishiyama, T.~Okabe, Y.~Sato, and I.~Sato.
\newblock Sensation-based photo cropping.
\newblock In {\em ACM Multimedia}, pages 669--672, 2009.

\bibitem{santella2006gaze}
A.~Santella, M.~Agrawala, D.~DeCarlo, D.~Salesin, and M.~Cohen.
\newblock Gaze-based interaction for semi-automatic photo cropping.
\newblock In {\em ACM SIGCHI}, pages 771--780, 2006.

\bibitem{simonyan2014very}
K.~Simonyan and A.~Zisserman.
\newblock Very deep convolutional networks for large-scale image recognition.
\newblock {\em arXiv preprint arXiv:1409.1556}, 2014.

\bibitem{stentiford2007attention}
F.~Stentiford.
\newblock Attention based auto image cropping.
\newblock In {\em ICVS Workshop on Computation Attention \& Applications},
  2007.

\bibitem{suh2003automatic}
B.~Suh, H.~Ling, B.~B. Bederson, and D.~W. Jacobs.
\newblock Automatic thumbnail cropping and its effectiveness.
\newblock In {\em ACM symposium on User interface software and technology},
  pages 95--104, 2003.

\bibitem{wang2017deep}
W.~Wang and J.~Shen.
\newblock Deep cropping via attention box prediction and aesthetics assessment.
\newblock In {\em ICCV}, 2017.

\bibitem{wang2018deep}
W.~Wang, J.~Shen, and H.~Ling.
\newblock A deep network solution for attention and aesthetics aware photo
  cropping.
\newblock {\em IEEE Transactions on Pattern Analysis and Machine Intelligence},
  2018.

\bibitem{wei2018good}
Z.~Wei, J.~Zhang, X.~Shen, Z.~Lin, R.~Mech, M.~Hoai, and D.~Samaras.
\newblock Good view hunting: Learning photo composition from dense view pairs.
\newblock In {\em CVPR}, pages 5437--5446, 2018.

\bibitem{wiki:xxx}
{Wikipedia contributors}.
\newblock Cropping (image) --- {Wikipedia}{,} the free encyclopedia.
\newblock
  \url{https://en.wikipedia.org/w/index.php?title=Cropping_(image)&oldid=847382681},
  2018.
\newblock [Online; accessed 10-July-2018].

\bibitem{wiki:rule-of-thirds}
{Wikipedia contributors}.
\newblock Rule of thirds --- {Wikipedia}{,} the free encyclopedia.
\newblock
  \url{https://en.wikipedia.org/w/index.php?title=Rule_of_thirds&oldid=852178012},
  2018.
\newblock [Online; accessed 31-July-2018].

\bibitem{yan2013learning}
J.~Yan, S.~Lin, S.~Bing~Kang, and X.~Tang.
\newblock Learning the change for automatic image cropping.
\newblock In {\em CVPR}, pages 971--978, 2013.

\bibitem{zhang2014weakly}
L.~Zhang, M.~Song, Y.~Yang, Q.~Zhao, C.~Zhao, and N.~Sebe.
\newblock Weakly supervised photo cropping.
\newblock {\em IEEE Transactions on Multimedia}, 16(1):94--107, 2014.

\bibitem{zhang2013probabilistic}
L.~Zhang, M.~Song, Q.~Zhao, X.~Liu, J.~Bu, and C.~Chen.
\newblock Probabilistic graphlet transfer for photo cropping.
\newblock {\em IEEE Transactions on Image Processing}, 22(2):802--815, 2013.

\bibitem{zhang2005auto}
M.~Zhang, L.~Zhang, Y.~Sun, L.~Feng, and W.~Ma.
\newblock Auto cropping for digital photographs.
\newblock In {\em ICME}, 2005.

\end{thebibliography}
}

\end{document}